\documentclass[eat,twocolumn]{jmlr}

\usepackage{longtable}

\usepackage{booktabs}
\usepackage[load-configurations=version-1]{siunitx} 

\newcommand{\ours}{Universal Healthcare Predictive Framework}
\newcommand{\oursarc}{UniHPF}
\theorembodyfont{\upshape}
\theoremheaderfont{\scshape}
\theorempostheader{:}
\theoremsep{\newline}

\usepackage{adjustbox}
\usepackage{booktabs, multirow}
\usepackage{graphicx}
\usepackage{paralist}
\usepackage{amsmath}
\usepackage{enumitem}
\usepackage{amsfonts}
\usepackage{soul}
\usepackage{multirow}
\usepackage{multicol}
\usepackage[compact]{titlesec}
\titlespacing{\section}{5pt}{2ex}{1ex}
\titlespacing{\subsection}{3pt}{1ex}{0ex}
\usepackage{bm,amsmath}
\usepackage{xspace}
\usepackage{graphicx}
\usepackage{color}

\usepackage{colortbl}
\definecolor{light_gray}{RGB}{170,170,170}

\usepackage[utf8]{inputenc} 
\usepackage[T1]{fontenc}    
\usepackage{url}            
\usepackage{booktabs}       
\usepackage{amsfonts}       
\usepackage{nicefrac}       
\usepackage{microtype}      
\usepackage{longtable,lscape}
\usepackage{array,multirow}
\usepackage{enumitem}

\usepackage{wrapfig}

\newcommand{\hide}[1]{}

\newcommand{\mb}{\mathbf{m}}

\newcommand{\pb}{\mathbf{p}}

\newcommand{\vb}{\mathbf{v}}

\newcommand{\yb}{\mathbf{y}}

\jmlrvolume{}
\firstpageno{1}

\jmlryear{2022}
\jmlrworkshop{Machine Learning for Health (ML4H) 2022}

\title[UniHPF]{UniHPF:Universal Healthcare Predictive Framework \titlebreak with Zero Domain Knowledge}

\author{
\Name{Kyunghoon Hur} \Email{pacesun@kaist.ac.kr} \\ 
\Name{Jungwoo Oh} \Email{ojw0123@kaist.ac.kr} \\
\Name{Junu Kim} \Email{kjune0322@kaist.ac.kr} \\
\Name{Jiyoun Kim} \Email{jiyoun.kim@kaist.ac.kr} \\
\Name{Min Jae Lee} \Email{mjbooo@kaist.ac.kr} \\
\Name{Eunbyeol Cho} \Email{eunbyeol.cho@kaist.ac.kr} \\
\addr KAIST \\
\Name{Seong-Eun Moon} \Email{seongeun.moon@navercorp.com} \\
\addr NAVER AI Lab \\
\Name{Younghak Kim} \Email{mdyhkim@amc.seoul.kr} \\
\addr Asan Medical Center, University of Ulsan College of Medicine \\
\Name{Edward Choi} \Email{edwardchoi@kaist.ac.kr}\\
\addr KAIST
}

\begin{document}

\maketitle

\begin{abstract}
Despite the abundance of Electronic Healthcare Records (EHR), its heterogeneity restricts the utilization of medical data in building predictive models.
To address this challenge, we propose Universal Healthcare Predictive Framework (UniHPF), which requires no medical domain knowledge and minimal pre-processing for multiple prediction tasks. 
Experimental results demonstrate that UniHPF is capable of building large-scale EHR models that can process any form of medical data from distinct EHR systems. 
We believe that our findings can provide helpful insights for further research on the multi-source learning of EHRs.
\end{abstract}

\section{Introduction} \label{sec:intro}

Patient medical records are accumulated regularly in the form of Electronic Health Records (EHR), enabling quality treatment based on patients' medical history. 
However, typical EHR datasets do not follow a single data format since each hospital stores EHR data according to their own needs.
Specifically, different EHR systems adopt different medical code standards (\textit{e.g., ICD-9, ICD-10, raw text}), and use distinct database schemas to store patient records~\citep{johnson2016mimic, johnson2021mimic, pollard2018eicu}.

Such heterogeneity is problematic because it acts as a barrier towards EHR model development.
In particular, when using patient clinical data, each hospital must employ its own data experts to rigorously pre-process EHR. Figure~\ref{fig:fig1} shows a typical framework for EHR-system-driven predictive models. In addition, discrepancies in medical codes and schemas prevent multiple healthcare organizations from conducting multi-source learning, such as further training a model that has been previously trained on a distinct EHR database (\textit{i.e., transfer learning}) or developing a model with EHR data pooled from multiple hospitals (\textit{i.e., pooled learning}).
 
Previous studies have attempted to overcome this dissimilarity in several ways. 
For instance, \citet{rajkomar2018scalable} used FHIR~\citep{mandel2016smart}, a type of Common Data Model (CDM) to manually standardize distinct EHR data into a single format.
In addition, DescEmb~\citep{hur2022unifying} aimed to overcome the heterogeneity of medical codes by utilizing clinical descriptions linked to each code, partially enabling multi-source learning.
Despite their progress, they still necessitate EHR system-specific healthcare expertise to select meaningful features.

In this work, we propose \ours{} (\oursarc).
Our framework presents a method for embedding any form of EHR systems for prediction tasks without requiring domain-knowledge-based pre-processing, such as medical code mapping and feature selection.
We believe that our findings can provide helpful insights for further research on the multi-source learning of EHR.


\begin{figure*}[ht]
\floatconts
    {fig:fig2}
    {\caption{
    Overview of \oursarc. On the top, a patient's series of medical events occur over time.
    Each medical event $\mathcal{M}_i$ is made up of event-related features $A_i^k$, including feature names and their values.
    These features, prepended with event type $e_i$, are converted to corresponding descriptions, and tokenized into a sequence of sub-words.
    Then an event encoder $f$ converts the sequence to an embedding $\mb_i$, which is passed to the event aggregator $g$, which then makes a prediction $\hat{\yb}$.
    } \vskip -25pt }
    {\includegraphics[width=0.6\textwidth]{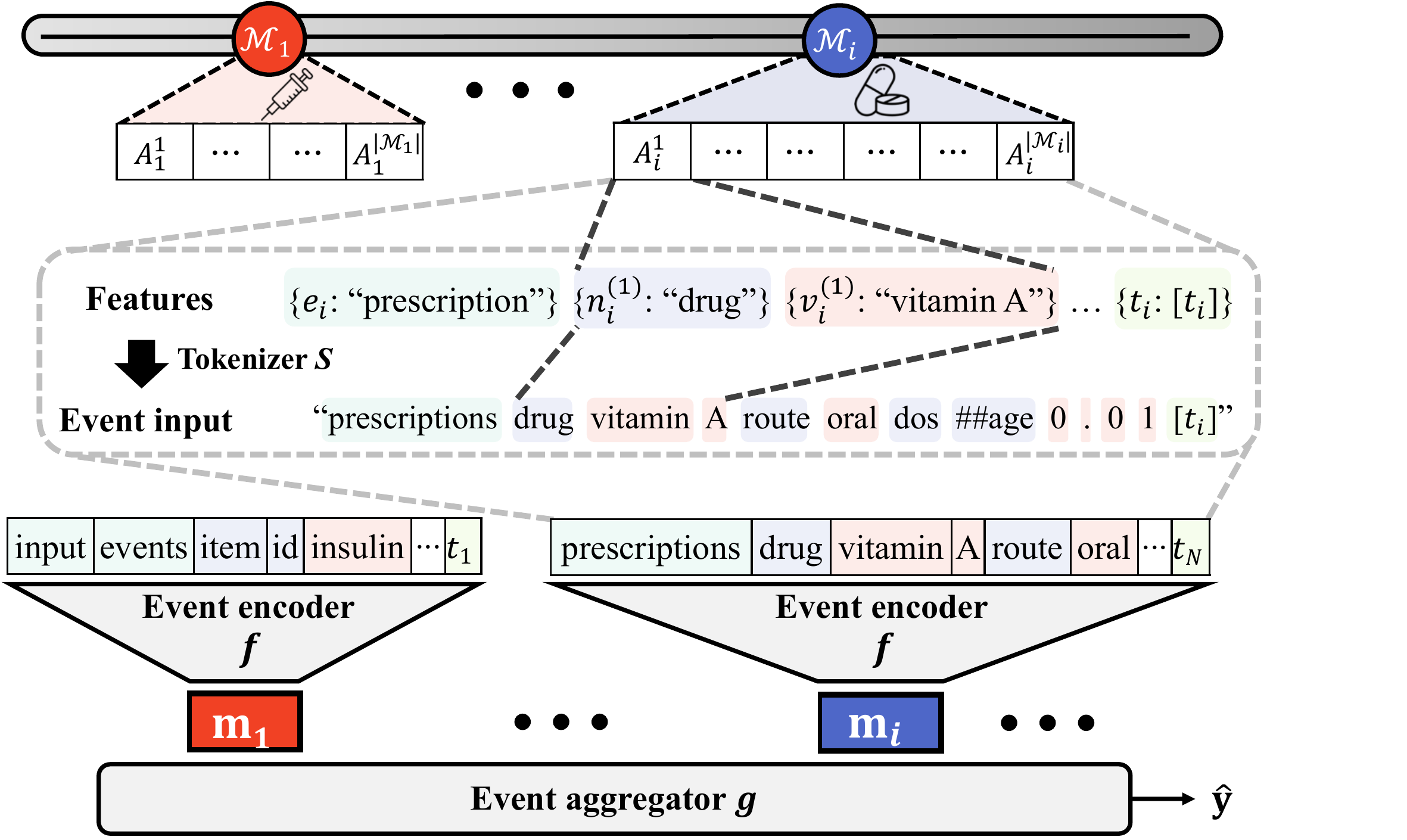}}
\end{figure*}

\section{Related work} \label{sec:related work}
\noindent{\textbf{Domain-knowledge-based predictive models.}}
Several studies on EHR-based prediction have attempted to fully utilize medical domain knowledge.
MIMIC-Extract~\citep{wang2020mimicextract} performs domain-knowledge-based feature engineering, such as grouping semantically similar concepts into clinical taxonomy.
Based on these heavily hand-crafted features, \citet{mcdermott2021a} proposed a benchmark for various healthcare predictive tasks.

\noindent{\textbf{Resolving heterogeneous EHR systems.}}
Researchers have been working on alternatives to overcome heterogeneity in EHR without CDM, which is considered as one of the main challenges in the modeling of medical data.
Meanwhile, AutoMap~\citep{wu2022automap} conducts medical code mapping via self-supervised learning with a predefined medical ontology which depends on EHR code systems.
In another study, instead of dealing with medical codes directly, DescEmb~\citep{hur2022unifying}exploits the text descriptions corresponding to each medical code.
However, DescEmb still requires the selection of features specific to each EHR system.

\section{Methodology} \label{sec:math}
\subsection{Structure of Electronic Health Records}

In typical EHR data, each patient $P$ can be represented as a sequence of medical events $[\mathcal{M}_1,\ldots,\mathcal{M}_N]$, where $N$ is the total number of events throughout the entire patient visit history. 
The i-th medical event of a patient $\mathcal{M}_i$ can be expressed as a set of event-associated features $\{A_i^{1},\ldots,A_i^{|\mathcal{M}_i|}\}$.
Each feature $A_i^{k}$ can be seen as a tuple of a feature name and its value $(n_i^{k}, v_i^{k}), n_i^k \in \mathcal{N}, v_i^k \in \mathcal{V}$, where $\mathcal{N}$ and $\mathcal{V}$ are each
a set of unique feature names 
(\textit{e.g.}, $\{$``drug name'', ``drug dosage'', $\ldots, \}$)
and feature values
(\textit{e.g.}, $\{$``vancomycin'', ``10.0'', $\ldots, \}$), respectively.

In addition, each medical event $\mathcal{M}_i$ has its corresponding event type $e_i \in \mathcal{E}$ which denotes the type of the event
(\textit{e.g.}, $\mathcal{E} = \{$``\textit{lab test}'', ``\textit{prescription}'', $\ldots, \}$).
Lastly, since the recorded time is also provided with $\mathcal{M}_i$, we can measure the time interval $t_i$ between $\mathcal{M}_i$ and $\mathcal{M}_{i+1}$.
\subsection{Universal Healthcare Predictive Framework}
In this section, we present \oursarc, a universal framework for EHR-based prediction, where
the overall architecture is depicted by Figure~\ref{fig:fig2}.

\noindent{\textbf{Text-based embedding.}}
A conventional EHR embedding method starts by assigning a unique embedding for each element in $\mathcal{V}$ via a linear map (\textit{i.e.}, lookup table) $f_{\mathcal{V}}$~\citep{choi2016multi, song2018attend, song2019medical, mcdermott2021a, rajkomar2018scalable}, so that $v_i^k$ can be converted to a vector $\vb_i^k \in \mathbb{R}^{d_v}$,
typically followed by pooling multiple feature values ($\vb_i^1, \vb_i^2, \ldots$) to obtain $\mb_i \in \mathbb{R}^{d_m}$, the embedding of $\mathcal{M}_i$

DescEmb~\citep{hur2022unifying} proposed to resolve this issue by suggesting a text-based embedding, where hospital-specific feature values are first converted to textual descriptions (\textit{e.g.}, ``401.9'' $\rightarrow$ ``unspecified essential hypertension'')
then a text encoder paired with a sub-word tokenizer is used to obtain $\mb_i$.
We extend the previous approach by applying the text-based embedding philosophy to event types $e_i$ and feature names $n_i^k$, in addition to feature values $v_i^k$, as follows:
\begin{equation*}
    \mb_i = f \Big( S(e_i), S(n_i^1), S(v_i^1), \ldots,
    [t_i] \Big) \label{eq:event_embedding}
\end{equation*}
where $S$ is a sub-word tokenizer, $f$ is an event encoder that takes a sequence of sub-word tokens and returns $\mb_i$, and $[t_i]$ is a special token for time intervals.

\noindent{\textbf{Employing the entire features of EHR.}\quad}
To develop a universal predictive framework,
we also must consider the \textit{schema heterogeneity}, namely each medical institution using different database schema.
When developing a conventional predictive model, medical domain experts are typically involved to define $\mathcal{M}'_i \subset \mathcal{M}_i$, a subset of task-specific features among $\mathcal{M}_i$ according to each EHR system.
Moreover, in multi-source learning, medical domain experts must select and match compatible features among distinct EHR systems.

To avoid this costly procedure, our framework exploits the entire features of medical events, effectively resolving the schema heterogeneity.
A formal comparison between previous and our approach to obtain $\mb_i$ is provided below:
\begin{flalign}
& \mbox{\textit{Conventional approach}:} \nonumber && \\
& \quad \mb_i = pool (\{ f_{\mathcal{V}} (v_i^k) \mid A_i^k \in \mathcal{M}'_i\} ) \nonumber && \\
& \mbox{\textit{DescEmb}:}  && \nonumber \\
& \quad \mb_i = f \Big( \{ S(v_i^k) \mid A_i^k \in \mathcal{M}'_i \} Big) \nonumber && \\
& \mbox{\textit{\oursarc}:}  && \nonumber \\
& \quad \mb_i = f \Big( S(e_i), \{ S(n_i^k), S(v_i^k) \mid A_i^k \in \mathcal{M}_i \} \Big) \nonumber
\end{flalign}

where \textit{pool} is typically implemented as concatenation or summation of the elements.
Note that we omitted the time interval in all equations to emphasize the fact that \oursarc{} differs from previous approaches in that it is the only approach to exploit all available information in a medical event: event type, all event names and all event values.
Therefore, \oursarc{} provides a general solution applicable to any EHR system with different schema, making it schema-agnostic without requiring medical domain knowledge.

\noindent{\textbf{Medical event aggregation.}}
To utilize the characteristics of EHR, where $P$ consists of a sequence of $\mathcal{M}_i$ and each $\mathcal{M}_i$ consists of a set of $A_i^k$,
we design a hierarchical model consisting of the event encoder $f$, and the event aggregator $g$.

After each $\mathcal{M}_i$ is converted to $\mb_i$ according to Eq.~\ref{eq:event_embedding}, we can obtain $\pb \in \mathbb{R}^{d_p}$, the vector representation of $P$ as follows:
\begin{equation*}
    \pb = g (\mb_1, \mb_2, \ldots, \mb_N)  \label{eq:hierarchical}
\end{equation*}
where $g$ is an embedding function that takes a sequence of event embeddings.

Note that it is possible to obtain $\pb$ by employing a flattened model architecture rather than a hierarchical one as follows:
\begin{align}
\pb &= h \Big( S(e_1), \{ S(n_1^k), S(v_1^k) \mid A_1^k \in \mathcal{M}_1 \}, [t_1], \nonumber \\
& \ldots, \nonumber \\
& S(e_N), \{ S(n_N^k), S(v_N^k) \mid A_N^k \in \mathcal{M}_N \}, [t_N] \Big) \nonumber \label{eq:flattened}
\end{align}
where sub-word tokens from all features of all medical events are passed to the sequence model $h$ at the same time.
We demonstrate that the hierarchical approach, which reflects the characteristics of EHR data, indeed outperforms the flattened approach.
These results are shown in Appendix~\ref{apd:hifl}.

\section{Experiments} \label{sec:exp}
\subsection{Experimental Settings}

We use baseline models to evaluate the feasibility of \oursarc{} for our objective, namely schema-agnostic EHR embedding without medical domain knowledge.
As there is no previous work, to our knowledge, that tackled exactly the same goal as ours, we modified well-known general-purpose EHR embedding frameworks(SAnD~\citep{song2018attend}, Rajkormar~\citep{rajkomar2018scalable}, DescEmb~\citep{hur2022unifying}).
In addition, all models were provided with both $n_i^k$ and $v_i^k$ for a fair comparison with \oursarc.
For datasets, three open source datasets (MIMIC-III, eICU, MIMIC-IV) were used. 
Datasets and baseline models details are provided in appendix \label{apd:data} and \label{apd:modeldetail}.


To evaluate our framework, we formulated seven prediction tasks: mortality(Mort), length of stay(LOS3, LOS7), readmission (Readm), final acuity (Fi\_ac), imminent discharge(Im\_disch), diagnosis (Dx)) following 
\citet{mcdermott2021a}.
All tasks are evaluated with the area under the precision recall curve (AUPRC).

\begin{figure*}[ht]
    \floatconts
    {fig:fig3}
    {\caption{Comparison of single domain prediction performance. The data source
    is represented on each row.
    The y-axis describes AUPRC
    and x-axis represents the models.
    The standard errors are provided by the error bars.
    } \vskip -25pt}
    {\includegraphics[width=1.0\linewidth]{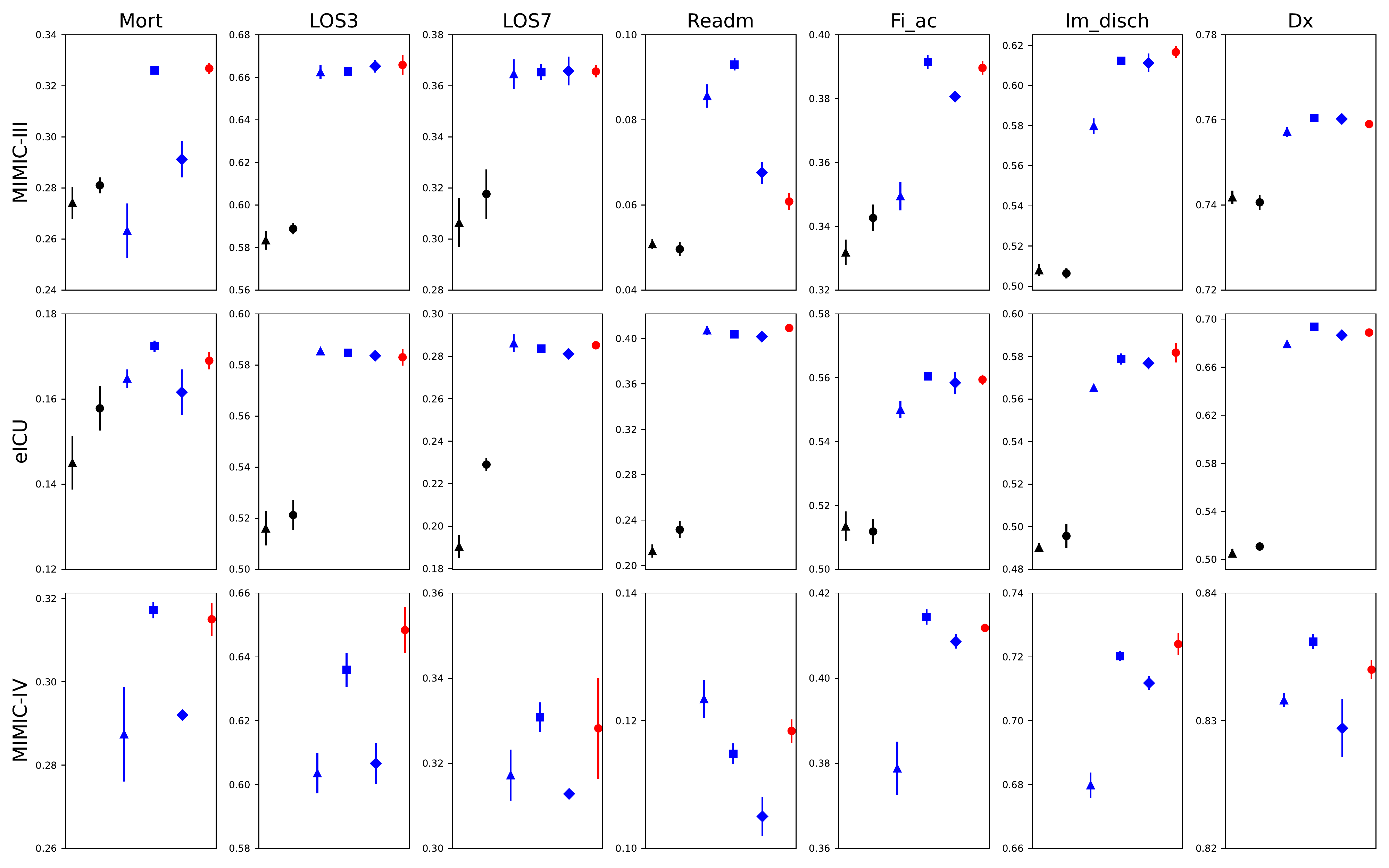}
    \includegraphics[width=0.8\linewidth]{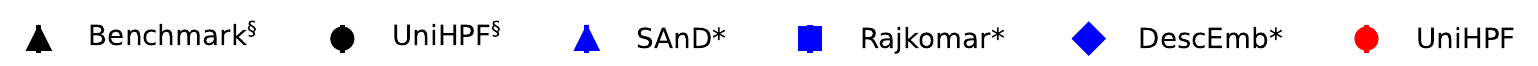}}
\end{figure*}

\begin{figure*}[ht]
    \floatconts
    {fig:fig4}
     {\caption{Pooled learning results. The data used for evaluation is represented on each row.
     The y-axis describes AUPRC and
    x-axis represents the models.
    The blue line separates models into
    conventional embedding models (left- SAnD*, Rajkomar*) and text-based embedding models (right- DescEmb*, UniHPF).
    Dot colors indicate the source datasets used for training.
    The $\star$ mark in each dot indicates p-value<0.05 from the t-test between single domain prediction (yellow dots) and pooled learning (other dots).
    Black arrows point from ``Single'' to ``MIMIC-III+MIMIC-IV+eICU''.
    } \vskip -25pt}
    {
    \includegraphics[width=1.0\linewidth]{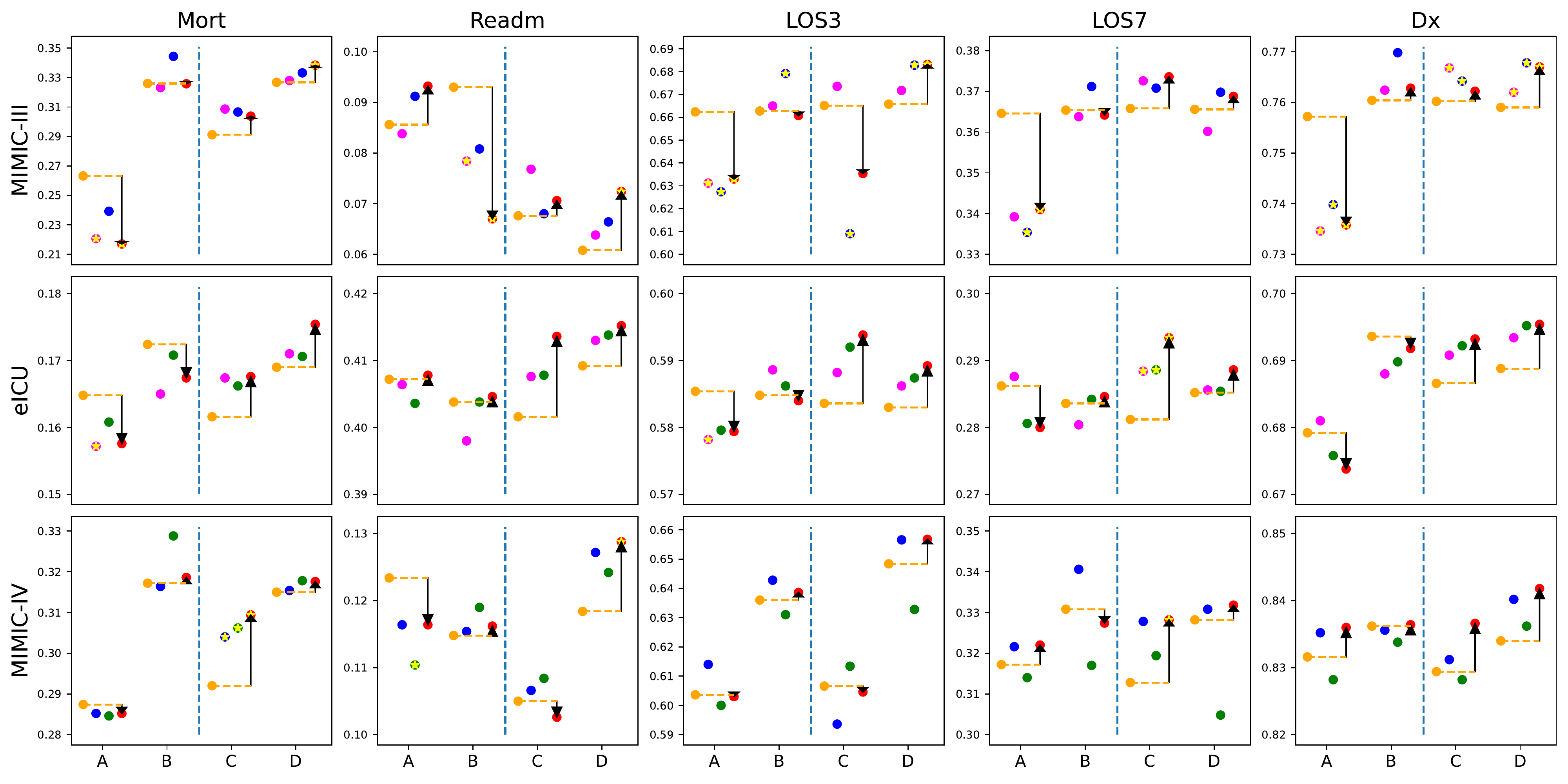}
    \includegraphics[width=0.8\linewidth]{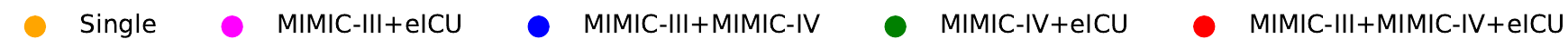}
    \includegraphics[width=0.7 \linewidth]{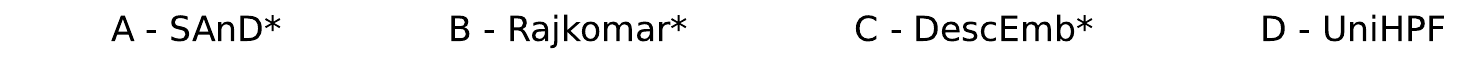}
    }
\end{figure*}

\subsection{Experimental Design}

\noindent{\textbf{Single domain prediction.}}
In single domain prediction, all models are trained on a single dataset's training set and tested on the same dataset's test set.
To provide credibility, we compare \oursarc{} with Benchmark~\citep{mcdermott2021a}.
Since Benchmark suggested an expert-designed feature-engineered prediction pipeline, comparing \oursarc{} with it can verify the effectiveness of our method, which does not involve any domain knowledge.
In this work we use a modified Benchmark$^\S$ to use only the lab test events, and compare it with a modified \oursarc{}$^\S$ that also only uses lab test events.

\noindent{\textbf{Multi-source learning.}}
To show the capability of our framework in multi-source learning, we set up the experiments on pooled learning and transfer learning scenario. 
For pooled leraning, we train the models on the pooled dataset from multiple sources, and evaluate them on each dataset's test set.


For transfer learning scenario, each model is first trained on a source dataset and then directly evaluated (\textit{i.e.,} zero-shot) or further trained (\textit{i.e.,} fine-tune) on a target dataset.
Here, we introduce two extra baselines that can be used to automatically map different code systems: AutoMap~\citep{wu2022automap} and MUSE~\citep{conneau2017word}.


\subsection{Single Domain Prediction}
The results of single domain prediction are shown in Figure~\ref{fig:fig3}.
First, we compare \oursarc{}$^\S$ with Benchmark$^\S$ to see how absence of domain knowledge affects prediction performance.
\oursarc{}$^\S$ generally shows higher performance than Benchmark$^\S$ in most prediction tasks.
This implies that it is possible to achieve better AUPRC without significant feature engineering.

Next, we compare all models that use lab tests, prescriptions, and input events.
\oursarc{} shows comparable prediction performance to models using domain knowledge and conventional embedding (SAnD*, Rajikomar*, DescEmb*) except the readmission tasks on MIMIC-III and MIMIC-IV.
In particular, a comparison between \oursarc{} and Rajkomar* suggests that it is unnecessary to assign unique embeddings for all feature names and values.
In addition, a comparison between \oursarc{} and DescEmb* demonstrates that applying medical domain knowledge to select a subset of meaningful features does not necessarily lead to greater performance than simply using all features.

\subsection{Multi-source Learning}
\label{sec:pooled}

The results of pooled learning are shown in Figure~\ref{fig:fig4}.
For text-based embedding models (DescEmb* and UniHPF), the results when training on the pooled dataset from all the three sources
consistently show higher performances than the single domain predictions 
In contrast, in the case of conventional embedding models (SAnD* and Rajkomar*),
the performances generally decrease.
We speculate that this result comes from the fact that MIMICs and eICU do not share any codes. Training conventional embedding models on this pooled dataset does nothing but expand the number of required embeddings for each feature name and value, which prevents the model from taking advantage of larger training data.

In addition, within text-based embedding models, \oursarc{} outperforms DescEmb* in most cases when all three data sources are pooled.
This result implies that \oursarc{} has a better capability to capture underlying semantics of distinct EHR sources than DescEmb*, by utilizing all available information in a medical event.

Next, the results of the transfer learning are presented in appendix Table~\ref{tab: transfer}. 
In the fine-tune and zero shot scenario, the results show that \oursarc{} outperforms the other methods in most cases.
In addition, compared to single domain prediction performance, we can see that \oursarc{} mostly benefits from the pre-trained source dataset.

\section{Conclusion} 
In this paper, we proposed a universal healthcare prediction framework, \oursarc{}, which enables multi-source learning by solving EHR heterogeneity of code and schema simultaneously, without medical domain knowledge or pre-processing.
The experimental results showed that \oursarc{} can act as a cornerstone for large-scale model training with multiple EHR sources.

\clearpage

\acks{This work was supported by Institute of Information \& Communications Technology Planning \& Evaluation (IITP) grant (No.2019-0-00075), Korea Medical Device Development Fund grant (Project Number: 1711138160, KMDF\_PR\_20200901\_0097), and the Korea Health Industry Development Institute (KHIDI) grant (No.HR21C0198), funded by the Korea government (MSIT, MOTIE, MOHW, MFDS).}
\bibliography{pmlr-sample}

\renewcommand{\arraystretch}{1.3}
\onecolumn

\appendix

\section{A typical EHR predictive model}
\begin{figure}[ht] 
\floatconts
    {fig:fig1}
    {\caption{ 
    A typical EHR predictive model framework involves domain-knowledge-based pre-processing for each medical center’s schema, which requires schema-specific and code system-specific feature engineering.
    }}
    {\includegraphics[width=0.7\linewidth]{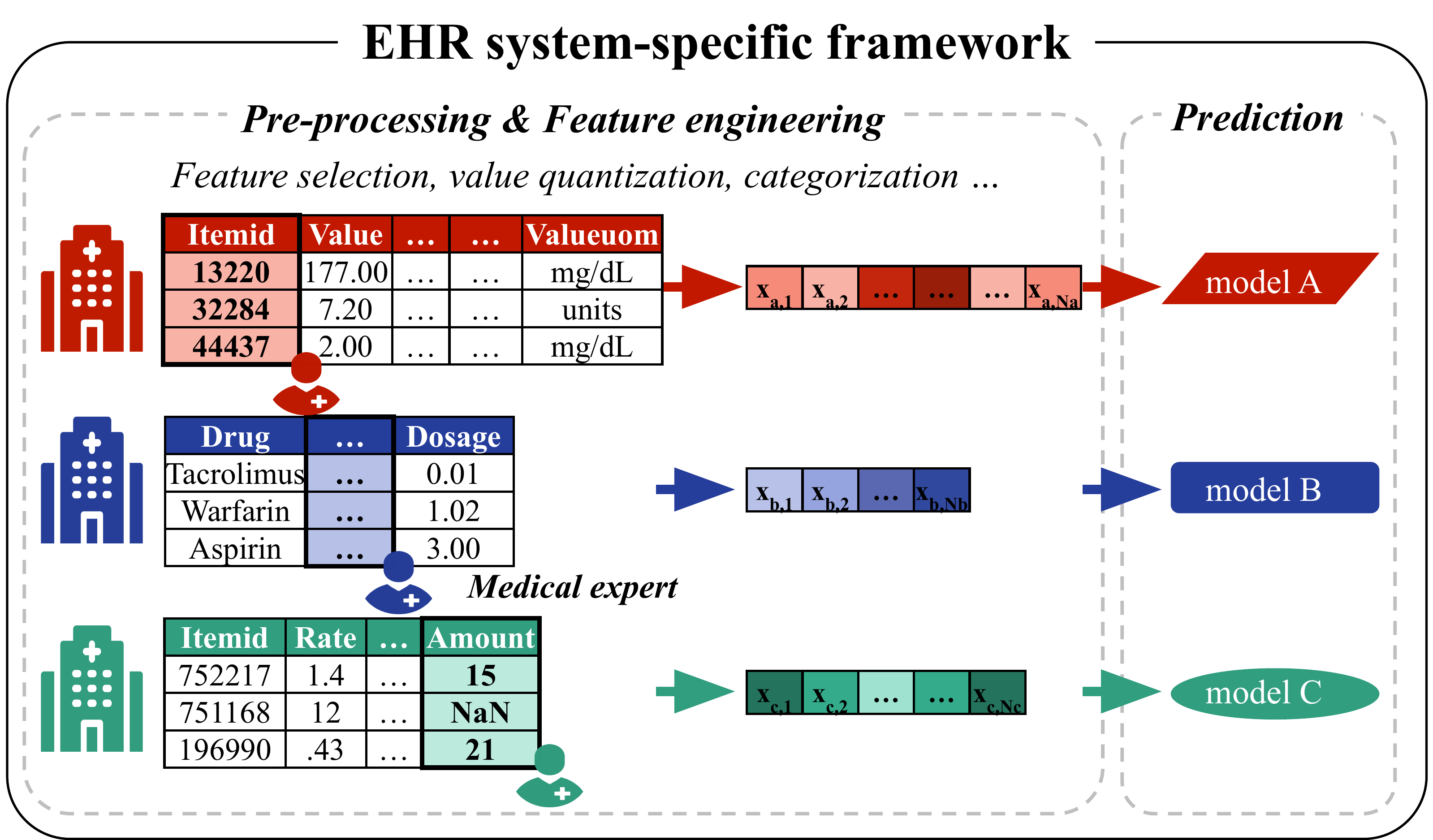}}
\end{figure}
\clearpage
\section{transfer learning}\label{apd:transfer}

\begin{table*}[ht]
\large{
\centering
\caption{AUPRCs of zero-shot test and fine-tune test results on five prediction tasks.
For both zero-shot and fine-tune, the best results are written in boldface for each row.
When fine-tuning, we additionally reported the performance difference with its single domain prediction.
}
\label{tab: transfer}
\adjustbox{width=1.1\textwidth,center}{
\begin{tabular}{ccccccc|ccccccc} \toprule
\multicolumn{7}{c}{\textbf{Zero-shot}} &\multicolumn{6}{c}{\textbf{Fine-tune}} \\
\hline
\multicolumn{13}{c}{MIMIC-III $\xrightarrow[]{}$ eICU} \\
\hline
 & SAnD* &AutoMap &MUSE &Rajkomar* & DescEmb* &\textbf{UniHPF}  &SAnD* &AutoMap &MUSE &Rajkomar* &DescEmb* &\textbf{UniHPF} \\
\hline
Mort &0.048 &0.044 &0.039 &0.035 &\textbf{0.135} &\textbf{0.135} &0.141(-0.024) &0.105(+0.039) &0.138(+0.081) &0.137(-0.035) &0.162(0) &\textbf{0.175(+0.006)} \\
LOS3 &0.452 &0.465 &0.463 &0.458 &0.503 &\textbf{0.507} &0.575(-0.01) &0.541(+0.014) &0.573(+0.052) &0.587(+0.002) &0.578(-0.006) &\textbf{0.589(+0.006)} \\
LOS7 &0.178 &0.186 &0.178 &0.192 &\textbf{0.264} &0.258 &0.259(-0.027) &$0.23(+0.02)^\dagger$ &0.263(+0.052) &$0.278(-0.006)^\dagger$ &0.28(-0.001) &\textbf{0.293(+0.008)} \\
Readm &0.166 &0.165 &0.154 &0.169 &0.320 &\textbf{0.333} &0.399(-0.008) &$0.25(-0.126)^\dagger$ &0.398(+0.038) &0.351(-0.053) &0.28(-0.122) &\textbf{0.415(+0.006)} \\
Dx &0.277 &$0.29^\dagger$ &$0.299^\dagger$ &0.289 &0.629 &\textbf{0.646} &0.672(-0.007) &0.435(-0.157) &0.675(+0.07) &\textbf{0.688(-0.006)} &0.682(-0.005) &\textbf{0.688(-0.001)} \\
\hline
\multicolumn{13}{c}{eICU $\xrightarrow[]{}$ MIMIC-III} \\
\hline
Mort &$0.048^\dagger$ &0.048 &0.048 &0.048 &0.238 &\textbf{0.245} &0.254(-0.009) &0.066(-0.054) &$0.227(+0.115)^\dagger$ &0.32(-0.006) &0.299(+0.008) &\textbf{0.333(+0.006)} \\
LOS3 &0.493 &0.482 &0.5 &0.492 &0.533 &\textbf{0.544} &0.643(-0.019) &0.612(+0.012) &0.652(+0.049) &\textbf{0.664(+0.001)} &0.659(-0.006) &0.663(-0.003) \\
LOS7 &0.219 &$0.213^\dagger$ &$0.204^\dagger$ &0.221 &0.292 &\textbf{0.308} &0.333(-0.032) &$0.29(-0.015)^\dagger$ &0.333(+0.036) &0.35(-0.015) &0.367(+0.001) &\textbf{0.379(+0.013)} \\
Readm &0.061 &0.063 &0.06 &\textbf{0.068} &0.049 &0.055 &0.076(-0.01) &0.065(-0.012) &0.077(-0.004) &\textbf{0.09(-0.003)} &0.066(-0.002) &$0.069(+0.008)^\dagger$ \\
Dx &0.533 &$0.523^\dagger$ &$0.521^\dagger$ &0.536 &0.639 &\textbf{0.647} &0.758(+0.001) &0.648(-0.054) &0.751(+0.049) &0.76(0) &0.754(-0.006) &\textbf{0.765(+0.006)} \\
\hline
\multicolumn{13}{c}{MIMIC-III $\xrightarrow[]{}$ MIMIC-IV} \\
\hline
Mort &0.018 &0.024 &0.024 &0.014 &0.222 &\textbf{0.228} &0.284(-0.003) &$0.151(-0.019)^\dagger$ &0.244(+0.076) &\textbf{0.326(+0.009)} &0.301(+0.009) &0.309(-0.006) \\
LOS3 &0.402 &0.411 &0.404 &0.389 &0.527 &\textbf{0.536} &0.592(-0.012) &$0.511(-0.013)^\dagger$ &0.582(+0.051) &0.614(-0.022) &0.607(+0.001) &\textbf{0.654(+0.006)} \\
LOS7 &0.172 &0.166 &0.164 &$0.171^\dagger$ &0.267 &\textbf{0.279} &0.283(-0.034) &0.233(-0.004) &0.295(+0.047) &0.288(-0.043) &0.319(+0.006) &\textbf{0.326(-0.002)} \\
Readm &0.08 &0.082 &0.082 &0.085 &0.082 &\textbf{0.095} &0.109(-0.014) &0.093(-0.016) &0.114(+0.007) &\textbf{0.131(+0.016)} &0.118(+0.013) &0.121(+0.003) \\
Dx &0.638 &$0.649^\dagger$ &$0.666^\dagger$ &0.624 &\textbf{0.778} &\textbf{0.788} &0.825(-0.007) &0.749(-0.032) &0.832(+0.051) &0.834(-0.002) &0.828(-0.001) &\textbf{0.841(+0.007)} \\
\hline
\multicolumn{13}{c}{MIMIC-IV $\xrightarrow[]{}$ MIMIC-III} \\
\hline
Mort &0.043 &0.037 &0.038 &0.044 &0.139 &\textbf{0.146} &0.284(+0.021) &$0.159(+0.039)^\dagger$ &0.255(+0.143) &\textbf{0.324(-0.002)} &0.302(+0.011) &\textbf{0.324(-0.003)} \\
LOS3 &0.494 &0.495 &0.508 &0.490 &0.572 &\textbf{0.586} &0.651(-0.011) &$0.598(-0.002)^\dagger$ &0.646(+0.043) &0.656(-0.007) &0.656(-0.009) &\textbf{0.66(-0.006)} \\
LOS7 &$0.187^\dagger$ &$0.192^\dagger$ &0.229 &0.196 &\textbf{0.303} &0.280 &$0.36(-0.005)^\dagger$ &$0.288(-0.017)^\dagger$ &0.328(+0.031) &\textbf{0.369(+0.004)} &$0.364(-0.002)^\dagger$ &0.367(+0.001) \\
Readm &0.049 &0.057 &0.059 &0.062 &0.062 &\textbf{0.070} &0.063(-0.023) &0.072(-0.005) &0.079(-0.002) &\textbf{0.095(+0.002)} &0.079(+0.011) &0.062(+0.001) \\
Dx &0.5 &$0.573^\dagger$ &$0.559^\dagger$ &0.495 &0.710 &\textbf{0.711} &\textbf{0.764(+0.007)} &$0.659(-0.043)^\dagger$ &0.754(+0.052) &0.75(-0.01) &$0.75(-0.01)^\dagger$ &0.761(+0.002) \\\midrule
\bottomrule
\end{tabular}}
\begin{flushleft}
    {\footnotesize †: standard deviation $>$ 0.02}
\end{flushleft}
}
\vspace{-7.5mm}
\end{table*}

The results of the transfer learning are presented in Table~\ref{tab: transfer}.
Specifically, in the case of zero-shot, we can see that the code-based embedding methods (SAnD*, AutoMap, MUSE, and Rajikomar*) consistently show inferior performances compared to the text-based embedding methods (DescEmb* and \oursarc), excluding the readmission task in MIMIC-III.
This again shows text-based embedding is more advisable to construct a unified healthcare framework than conventional embedding
In addition, \oursarc{} generally exhibits the best performance for prediction tasks across various transfer scenarios.
This also shows that using all available information (\oursarc) is more helpful for learning the semantics of medical code descriptions rather than selecting and matching specific features (DescEmb*), as also mentioned in Sec.~\ref{sec:pooled}.

\clearpage

\section{Dataset details}\label{apd:data}
We draw on three publicly available datasets; MIMIC-III~\citep{johnson2016mimic}, MIMIC-IV~\citep{johnson2021mimic}, and eICU~\citep{pollard2018eicu}.

All three datasets contain patient medical events including lab tests, prescriptions, and input events (\textit{e.g.}, drug injection), each event marked with timestamps.
MIMIC-III and MIMIC-IV share the same code system with similar schemas, whereas eICU has a completely distinct code system and schema.

To ensure reliable experiments and analysis, we split the dataset into training, validation and test sets according to 8:1:1 ratio in a stratified manner for each target label.
All experiments were conducted with five random seeds and we report the mean performance.

More information about how datasets were created is provided in this section.
\begin{itemize}
\item The MIMIC-III database consists of clinical data of over 40,000 patients admitted to intensive care units (ICU) at the Beth Israel Deaconess Medical Center from 2001 to 2012.
\item MIMIC-IV is an updated version of MIMIC-III that includes new sources of data, admission date shifting, and extended period of records collected from 2008 to 2019.
\item eICU consists of ICU records from multiple US-based hospitals, totaling up to 140,000 unique patients admitted between 2014 and 2015.
\end{itemize}
For applying \oursarc{} to any EHR datasets, only two pre-processing steps are necessary, which do not involve any domain knowledge.
First, we drop features whose values only consist of integers.
This automatically leads to using all continuous-valued features (\textit{e.g.}, lab test result) and textual features (\textit{e.g.}, lab test name), while features such as patient ID are removed.
Second, we split numeric values digit-by-digit and assign a special token for each digit place, namely \textit{digit place embedding}, which was first introduced in DescEmb~\citep{hur2022unifying}.

\subsection{Table selection}
\begin{table}[ht]
    \centering
    \scriptsize
    \caption{\label{tab:filesource} File names for each data sources}
    \begin{tabular}{|c|c|c|c|l}
    \cline{1-4}
    \textbf{}             & \textbf{MIMIC-III} & \textbf{eICU}    & \textbf{MIMIC-IV} &  \\ \cline{1-4}
    \textbf{Lab test}     & LABEVENTS.csv      & lab.csv          & labevents.csv     &  \\ \cline{1-4}
    \textbf{Prescription} & PRESCRIPTIONS.csv  & medication.csv   & prescriptions.csv &  \\ \cline{1-4}
    \textbf{Infusion}     & INPUTEVENTS.csv    & infusionDrug.csv & inputevents.csv   &  \\ \cline{1-4}
    \end{tabular}
\end{table}
For each patient, three sources with different “event types” (lab tests, prescription, and infusion) are preprocessed as input for a predictive model. Table ~\ref{tab:filesource} lists csv filenames with each event type.
Note that MIMIC-III files ’INPUTEVENTS MV’ and ’INPUTEVENTS CV’, are merged and named as INPUTEVENTS.
File names for each data sources and tables are described below.

\subsection{Patient cohort setup}
\begin{table}[ht]
    \caption{\label{tab:tabstat} Prediction datasets summary statistics}
    \centering
    \scriptsize
    \begin{tabular}{l l l l}
    \toprule
    Statistic          &  MIMIC-III  &  eICU  &  MIMIC-IV    \\
    \hline
    \midrule
     No. of Observations   &  38040     &    98904   &    65511  \\
     No. of ICU stays      &  38040     &     98904   &   65511   \\
     No. of Unique codes   &  10385      &    6302      &   9565    \\
     No. of Unique subwords   &  2235      &    1585     &   2724    \\
    Mean No. of events per sample  &  98.47    &    538.89   &   116.32  \\
    Mean of code length per event   &  18.13     &    16.82      &  21.03    \\
    Mean of subword length per event &  50.28   &    47.91      &  69.75    \\

    \bottomrule
    \end{tabular}
\end{table}
For the sake of comparability, we built patients cohorts from MIMIC-III, MIMIC-IV and eICU databases based on the following criteria: patients over the age of 18 years who remained in the ICU for over 24 hours.
Then, we exclusively consider the first ICU stay during a single hospital stay, and remove any ICU stays with fewer than five medical events.
Within each ICU stay, we restrict our samples to the first 12 hours of data, and remove features that occur fewer than five times in the entire dataset.
Lastly, we eliminate events with lower frequency of main columns (drug name, ITEMID, … ). Cohort summary is described in Table ~\ref{tab:tabstat}

\subsection{Convert EHR table to input sequence}
Here, we will explain our pre-process algorithm which enables us to deal with any EHR table, converting them into the same input configuration for UniHPF. The process explanation is represented below.

\begin{enumerate}[topsep=0pt]
\item First, replace code features to description if the definition table exists in the EHR source set, which the definition table has features as key and description as value. (e.g. MIMIC-III DITEMS.csv)
\item Remove columns whose data type is integer except columns which have categorical values (e.g. number of unique features <50).
\item Select the associated timestamp column which is most relevant to the point of occurrence and drop the other timestamp columns.
\item  Convert all features as string type and tokenize them with “bio-clinical-bert” tokenizer except associated timestamp columns.
\item  For numeric values in feature, split them digit by digit before being tokenized and apply digit-place embedding (DPE) following the value embedding method from DescEmb, which assigns a special token for each digit place.
\item  Descriptions corresponding to each event are listed in the order of event type, feature name, and feature value.
\item  For time stamps, the time interval between the corresponding event and the next event is used as the time feature. 8. At this time, we follow Rajikomar method to deal with continuous values, quantizing them into discretized features. So, the time interval is bucketed into 20 separations within the entire time interval and converted to special tokens.
\item Next, create a class of type token corresponding to the event type, feature name, and feature value.
\item  Lists events in order based on timestamp. Note that these type tokens are used as indicating each sub-word token type (event type, feature name, feature value).
This type token sequence is added to event input with sinusoidal positional embedding.
\item  Finally, prepare an input dataset with a shape as (N, S, W), where N is a number of icu stay, S is maximum length of events, and W is maximum sub-word length for each event.
\end{enumerate}

\subsection{Datasets preparation for each model}

\begin{enumerate}[topsep=0pt]
\item Feature selection
\begin{itemize}[topsep=0pt]
\item We prepare two versions of the dataset, feature selection version and without feature selection version (using Entire EHR).
\item This was to compare the case with and without the conventional feature selection process, and in the case of SAnD and DescEmb, the feature selected dataset is used.
\item Feature selection criteria follows DescEmb, which are using information corresponding to medical code, numerical value, unit of measurement.
\end{itemize}
\item Conventional embedding method
\begin{itemize}[topsep=0pt]
\item Each feature is coded based on unique text.
\item Before converting feature text into unique code, continuous values are buckettized after being grouped by each ITEMID.
\item For categorical features, preprocessing is performed separately on categorical code.
\item Feature names (columns) are also converted as codes.
\end{itemize}
\item Flattened structure
\begin{itemize}[topsep=0pt]
\item  The hierarchical form (N, S, W) of input data is reshaped into the shape of (B, SxW).
\item  After removing the pad in each W, flattened input shape is changed to (N, S*) where S* indicates flattened input without pad.
\item  SAnD* used this flattened dataset as input.
\item  The ablation study results for flatten and hierarchical are below.
\end{itemize}

\end{enumerate}

\section{Prediction task}\label{apd:prediction}
\begin{table}[ht]
\centering
\caption{\label{tab:labeling} Detailed label definition in the code}
\scriptsize
\begin{tabular}{|l|l|}
\hline
\multirow{2}{*}{\textbf{Target}} & \multirow{2}{*}{\textbf{MIMIC}}                                                                                                 \\
                                 &                                                                                                                                 \\ \hline
\multirow{2}{*}{Mortality}       & \multirow{2}{*}{'unitDischargeStatus'=='Expired' and (Timegap\textless   discharge time -INTIME \textless prediction window)}   \\
                                 &                                                                                                                                 \\ \hline
\multirow{2}{*}{LOS3}            & \multirow{2}{*}{LOS \textgreater{}3}                                                                                            \\
                                 &                                                                                                                                 \\ \hline
\multirow{2}{*}{LOS7}            & \multirow{2}{*}{LOS \textgreater{}7}                                                                                            \\
                                 &                                                                                                                                 \\ \hline
\multirow{2}{*}{Readm}           & \multirow{2}{*}{Count(‘ICUSTAY ID’) \textgreater{}1}                                                                            \\
                                 &                                                                                                                                 \\ \hline
\multirow{2}{*}{Fi\_ac}          & \multirow{2}{*}{class('hospitaldischargelocation') and   (Timegap\textless discharge time -INTIME \textless prediction window)} \\
                                 &                                                                                                                                 \\ \hline
\multirow{2}{*}{Im\_disch}       & \multirow{2}{*}{class('hospitaldischargelocation')}                                                                             \\
                                 &                                                                                                                                 \\ \hline
\multirow{2}{*}{Dx}              & \multirow{2}{*}{ICD9 CODE-LONG TITLE (MIMIC-III) ICD10 CODE-LONG   TITLE (MIMIC-IV)}                                            \\
                                 &                                                                                                                                 \\ \hline
                                 & \textbf{eICU}                                                                                                                   \\ \hline
\multirow{2}{*}{Mortality}       & \multirow{2}{*}{(DOD HOSP not null) and (Timegap\textless discharge   time -INTIME \textless prediction window)}                \\
                                 &                                                                                                                                 \\ \hline
\multirow{2}{*}{LOS3}            & \multirow{2}{*}{'unitDischargeOffset' \textgreater 32460}                                                                       \\
                                 &                                                                                                                                 \\ \hline
\multirow{2}{*}{LOS7}            & \multirow{2}{*}{'unitDischargeOffset' \textgreater 72460}                                                                       \\
                                 &                                                                                                                                 \\ \hline
\multirow{2}{*}{Readm}           & \multirow{2}{*}{Count('patientUnitStayID') \textgreater 1}                                                                      \\
                                 &                                                                                                                                 \\ \hline
\multirow{2}{*}{Fi\_ac}          & \multirow{2}{*}{class(DISCHARGE\_LOCATION) and   (Timegap\textless discharge time -INTIME \textless prediction window)}         \\
                                 &                                                                                                                                 \\ \hline
\multirow{2}{*}{Im\_disch}       & \multirow{2}{*}{class(DISCHARGE\_LOCATION)}                                                                                     \\
                                 &                                                                                                                                 \\ \hline
\multirow{2}{*}{Dx}              & \multirow{2}{*}{set('diagnosisstring') per 1 ICU}                                                                               \\
                                 &                                                                                                                                 \\ \hline
\end{tabular}
\end{table}
Following \citet{mcdermott2021a}, prediction tasks are well defined.
Medical event information from ICU admission to 12 hours duration is used, and TimeGAP is given 12 hours for all tasks.
The rolling type task (mortality, imminent discharge) is applied only for the first rolling point(similar to static type task), and the prediction window was given at 48hr.
In the case of diagnosis, we tried to group CCS into 18 diagnosis classes based on CCS ontology. MIMIC-III, MIMIC-IV and eICU used “Diagnosis.csv”, “diagnoses icd.csv” and “diagnosis.csv” respectively.
Detailed label definition in the Table ~\ref{tab:labeling}.

\clearpage
\section{Implementation details}\label{apd:implementation detail}

\subsection{Baseline model details}\label{apd:modeldetail}
\begin{compactitem}
\item SAnD*: This uses the conventional embedding, selected features $\mathcal{M}'_i$, and the flattened architecture, similar in spirit to SAnD~\citep{song2018attend}.
Note that feature embeddings from all medical events $[\mathcal{M}_1, \ldots, \mathcal{M}_N]$ are directly fed to the sequence encoder $h$ instead of being pooled to obtain individual $\mb_i$.
\item Rajkomar*: This uses the conventional embedding, entire features $\mathcal{M}_i$, and the hierarchical approach, similar in spirit to \cite{rajkomar2018scalable} except the CDM standardization.
Note that feature embeddings from each $\mathcal{M}_i$ are fed to $f$ to obtain individual $\mb_i$, which is fed to $g$.
\item DescEmb*: This uses the text-based embedding, selected features $\mathcal{M}'_i$, and the hierarchical approach, similar in spirit to DescEmb~\citep{hur2022unifying}.
\end{compactitem}
For a fair comparison, $f$ and $g$ were both implemented with a randomly initialized 2-layer Transformer encoder, and $h$ a 4-layer Transformer encoder, making all models equivalent in terms of number of trainable parameters.
Further implementation details including the list of selected features $\mathcal{M}'_i$ For example, from the prescription event, we chose essential features such as drug name, drug volume, unit of measurement among all available features.

\subsection{Model architecture for each model}
\begin{table}[ht]
\centering
\caption{\label{tab:modelsummary} Comparison models on detail.}
\scriptsize
\begin{tabular}{|c|c|c|c|l}
\cline{1-4}
\textbf{Model}      & \textbf{Embedding} & \textbf{Feature} & \textbf{Structure}                            &  \\ \cline{1-4}
\textbf{UniHPF}     & Text based         & Entire           & Hierarchical (Transformer 2 layer + 2 layer ) &  \\ \cline{1-4}
\textbf{DescEmb*}   & Text based         & Selected         & Hierarchical (Transformer 2 layer + 2 layer ) &  \\ \cline{1-4}
\textbf{Rajikomar*} & Code based         & Entire           & Hierarchical (Transformer 2 layer + 2 layer ) &  \\ \cline{1-4}
\textbf{SAnD*}      & Code based         & Selected         & Flatten (Transformer 4 layer )                &  \\ \cline{1-4}
\end{tabular}
\end{table}

UniHPF and baseline models can be distinguished in the view of embedding method, feature usage and model structure. The comparison of models is in Table \ref{tab:modelsummary}

\subsection{Hyperparameters}
We searched for the ideal set of hyperparameters for each case for more than 72 hours. We found that the hyperparameters had little impact on the outcome.
We combined one set of hyperparameters for all cases to make the experiment more straightforward without significantly degrading the performance of each individual model.
The final results show a dropout of 0.3, a predictive model's embedding dimension being 128 and a learning rate of 1e-4.
\subsection{Computational resources}
\begin{table}[ht]
\centering
\caption{\label{tab:memory} VRAM usage of each model and parameters}
\scriptsize
\begin{tabular}{|l|l|l|l|l|}
\hline
\textbf{}                               & \textbf{SAnD*} & \textbf{DescEmb*} & \textbf{Rajikomar*} & \textbf{UniHPF} \\ \hline
\textbf{Memory}                         & 8.9GB          & 65.1GB            & 35.4GB              & 78.8GB * 2GPU   \\ \hline
\textbf{Total Parameters}               & 1746945        & 4414465           & 1970561             & 4414465         \\ \hline
\textbf{Parameters w/o embedding layer} & 1056897        & 396929            & 396929              & 396929          \\ \hline
\end{tabular}
\end{table}
VRAM memory usage was observed when the batch size was 128 based on the LOS3 prediction task, which is a binary classification in single domain training.
In the case of the flattened model SAnD, the input sequence length is 8192, but the VRAM usage is much reduced by using a performer which is efficient transformer ~\citep{choromanski2020rethinking}. Each model VRAM usage information is in Table \ref{tab:memory} 

\subsection{Training details}
We splitted train set, valid set, test set with 9:1:1 ratio and split is stratified for each prediction task.
Training model is saved for best prediction performance at valid testset and early stopping with 10 epoch patience is applied.
For pooled learning, a model with pooled datasets is trained and evaluated for a valid set of each dataset. Test best performance model on each dataset.
For transfer learning, a single domain trained model with source datasets is loaded and used for zero-shot learning or further fine-tuning on target datasets.

\section{Hierarchical vs Flatten Model}\label{apd:hifl}
\begin{table}[ht]
\centering
\scriptsize
\caption{\label{tab:hifl} Ablation study for hierarchical versus flattened model}
\begin{tabular}{|cc|c|c|c|c|}
\hline
\multicolumn{2}{|c|}{\textbf{MIMIC-III}}                                      & SAnD*(fl) & Rajikomar*(hi) & DescEmb*(hi) & UniHPF(hi) \\ \hline
\multicolumn{1}{|c|}{\multirow{2}{*}{\textbf{Readm}}} & \textbf{Flatten}      & 0.086     & 0.094          & 0.085        & 0.078      \\ \cline{2-6} 
\multicolumn{1}{|c|}{}                                & \textbf{Hierarchical} & 0.084     & 0.093          & 0.068        & 0.061      \\ \hline
\multicolumn{1}{|c|}{\multirow{2}{*}{Mort}}           & \textbf{Flatten}      & 0.263     & 0.316          & 0.277        & 0.29       \\ \cline{2-6} 
\multicolumn{1}{|c|}{}                                & \textbf{Hierarchical} & 0.29      & 0.326          & 0.291        & 0.327      \\ \hline
\multicolumn{1}{|c|}{\multirow{2}{*}{LOS3}}           & \textbf{Flatten}      & 0.662     & 0.663          & 0.657        & 0.661      \\ \cline{2-6} 
\multicolumn{1}{|c|}{}                                & \textbf{Hierarchical} & 0.662     & 0.663          & 0.665        & 0.666      \\ \hline
\multicolumn{1}{|c|}{\multirow{2}{*}{LOS7}}           & \textbf{Flatten}      & 0.365     & 0.359          & 0.364        & 0.358      \\ \cline{2-6} 
\multicolumn{1}{|c|}{}                                & \textbf{Hierarchical} & 0.364     & 0.365          & 0.366        & 0.366      \\ \hline
\multicolumn{2}{|c|}{\textbf{eICU}}                                           & SAnD*(fl) & Rajikomar*(hi) & DescEmb*(hi) & UniHPF(hi) \\ \hline
\multicolumn{1}{|c|}{\multirow{2}{*}{\textbf{Readm}}} & \textbf{Flatten}      & 0.407     & 0.403          & 0.396        & 0.401      \\ \cline{2-6} 
\multicolumn{1}{|c|}{}                                & \textbf{Hierarchical} & 0.403     & 0.404          & 0.402        & 0.409      \\ \hline
\multicolumn{1}{|c|}{\multirow{2}{*}{Mort}}           & \textbf{Flatten}      & 0.165     & 0.169          & 0.135        & 0.148      \\ \cline{2-6} 
\multicolumn{1}{|c|}{}                                & \textbf{Hierarchical} & 0.164     & 0.172          & 0.162        & 0.169      \\ \hline
\multicolumn{1}{|c|}{\multirow{2}{*}{LOS3}}           & \textbf{Flatten}      & 0.585     & 0.588          & 0.574        & 0.57       \\ \cline{2-6} 
\multicolumn{1}{|c|}{}                                & \textbf{Hierarchical} & 0.584     & 0.585          & 0.577        & 0.583      \\ \hline
\multicolumn{1}{|c|}{\multirow{2}{*}{LOS7}}           & \textbf{Flatten}      & 0.286     & 0.289          & 0.276        & 0.272      \\ \cline{2-6} 
\multicolumn{1}{|c|}{}                                & \textbf{Hierarchical} & 0.282     & 0.284          & 0.281        & 0.285      \\ \hline
\multicolumn{2}{|c|}{\textbf{MIMIC-IV}}                                       & SAnD*(fl) & Rajikomar*(hi) & DescEmb*(hi) & UniHPF(hi) \\ \hline
\multicolumn{1}{|c|}{\multirow{2}{*}{\textbf{Readm}}} & \textbf{Flatten}      & 0.123     & 0.116          & 0.117        & 0.12       \\ \cline{2-6} 
\multicolumn{1}{|c|}{}                                & \textbf{Hierarchical} & 0.12      & 0.115          & 0.105        & 0.118      \\ \hline
\multicolumn{1}{|c|}{\multirow{2}{*}{Mort}}           & \textbf{Flatten}      & 0.287     & 0.318          & 0.275        & 0.294      \\ \cline{2-6} 
\multicolumn{1}{|c|}{}                                & \textbf{Hierarchical} & 0.311     & 0.317          & 0.292        & 0.315      \\ \hline
\multicolumn{1}{|c|}{\multirow{2}{*}{LOS3}}           & \textbf{Flatten}      & 0.604     & 0.624          & 0.592        & 0.609      \\ \cline{2-6} 
\multicolumn{1}{|c|}{}                                & \textbf{Hierarchical} & 0.619     & 0.636          & 0.607        & 0.648      \\ \hline
\multicolumn{1}{|c|}{\multirow{2}{*}{LOS7}}           & \textbf{Flatten}      & 0.317     & 0.335          & 0.305        & 0.317      \\ \cline{2-6} 
\multicolumn{1}{|c|}{}                                & \textbf{Hierarchical} & 0.313     & 0.331          & 0.313        & 0.328      \\ \hline
\end{tabular}
\end{table}
For giving the same information between hierarchical and flatten models, We restricted the number of events for each sample. Due to the computation resource limitation, flattened models use 8192 as maximum sequence length and corresponding number of events is used on hierarchical model input.
Experiments were conducted to compare the structures of each model in flatten and hierarchical cases. The origin structure of each model is displayed in parenthesis. In most cases, hierarchical performance is higher than flatten structure regardless of model type.
This confirmed that embedding and aggregation of time-series EHR in event units is a more favorable condition for the model.
The result of ablation study is in Talbe \ref{tab:hifl}

\section{Pre-training}\label{apd:pretrain}
\begin{table}[ht]
\centering
\caption{\label{tab:pretrain} Pre-training results}
\scriptsize

\begin{tabular}{|c|c|ccccc|}
\hline
\textbf{}                           & \textbf{Pretraining Dataset} & \multicolumn{5}{c|}{\textbf{MIMIC-III + MIMIC-IV + eICU}}                                                                                                                           \\ \hline
                                    & \textbf{Model}               & \multicolumn{2}{c|}{\textbf{Hierarchical}}                                        & \multicolumn{3}{c|}{\textbf{Flatten}}                                                           \\ \hline
\textbf{Eval Datasets}              & \textbf{Task}                & \multicolumn{1}{c|}{\textbf{UniHPF (hi)}} & \multicolumn{1}{c|}{\textbf{Wav2Vec}} & \multicolumn{1}{c|}{\textbf{UniHPF(fl)}} & \multicolumn{1}{c|}{\textbf{MLM}} & \textbf{SPANMLM} \\ \hline
\multirow{7}{*}{\textbf{MIMIC-III}} & Mort                         & \multicolumn{1}{c|}{0.327}                & \multicolumn{1}{c|}{0.325}            & \multicolumn{1}{c|}{0.290}               & \multicolumn{1}{c|}{0.291}        & 0.293            \\ \cline{2-7} 
                                    & LOS3                         & \multicolumn{1}{c|}{0.666}                & \multicolumn{1}{c|}{0.663}            & \multicolumn{1}{c|}{0.661}               & \multicolumn{1}{c|}{0.664}        & 0.663            \\ \cline{2-7} 
                                    & LOS7                         & \multicolumn{1}{c|}{0.366}                & \multicolumn{1}{c|}{0.364}            & \multicolumn{1}{c|}{0.358}               & \multicolumn{1}{c|}{0.358}        & 0.357            \\ \cline{2-7} 
                                    & Readm                        & \multicolumn{1}{c|}{0.061}                & \multicolumn{1}{c|}{0.601}            & \multicolumn{1}{c|}{0.078}               & \multicolumn{1}{c|}{0.068}        & 0.073            \\ \cline{2-7} 
                                    & Fi\_ac                       & \multicolumn{1}{c|}{0.617}                & \multicolumn{1}{c|}{0.616}            & \multicolumn{1}{c|}{0.600}               & \multicolumn{1}{c|}{0.606}        & 0.601            \\ \cline{2-7} 
                                    & Im\_disch                    & \multicolumn{1}{c|}{0.390}                & \multicolumn{1}{c|}{0.389}            & \multicolumn{1}{c|}{0.375}               & \multicolumn{1}{c|}{0.379}        & 0.379            \\ \cline{2-7} 
                                    & Dx                           & \multicolumn{1}{c|}{0.759}                & \multicolumn{1}{c|}{0.761}            & \multicolumn{1}{c|}{0.753}               & \multicolumn{1}{c|}{0.756}        & 0.755            \\ \hline
\multirow{7}{*}{\textbf{eICU}}      & Mort                         & \multicolumn{1}{c|}{0.169}                & \multicolumn{1}{c|}{0.167}            & \multicolumn{1}{c|}{0.148}               & \multicolumn{1}{c|}{0.150}        & 0.151            \\ \cline{2-7} 
                                    & LOS3                         & \multicolumn{1}{c|}{0.583}                & \multicolumn{1}{c|}{0.579}            & \multicolumn{1}{c|}{0.570}               & \multicolumn{1}{c|}{0.574}        & 0.572            \\ \cline{2-7} 
                                    & LOS7                         & \multicolumn{1}{c|}{0.285}                & \multicolumn{1}{c|}{0.281}            & \multicolumn{1}{c|}{0.272}               & \multicolumn{1}{c|}{0.278}        & 0.278            \\ \cline{2-7} 
                                    & Readm                        & \multicolumn{1}{c|}{0.409}                & \multicolumn{1}{c|}{0.404}            & \multicolumn{1}{c|}{0.401}               & \multicolumn{1}{c|}{0.402}        & 0.400            \\ \cline{2-7} 
                                    & Fi\_ac                       & \multicolumn{1}{c|}{0.582}                & \multicolumn{1}{c|}{0.574}            & \multicolumn{1}{c|}{0.560}               & \multicolumn{1}{c|}{0.558}        & 0.561            \\ \cline{2-7} 
                                    & Im\_disch                    & \multicolumn{1}{c|}{0.559}                & \multicolumn{1}{c|}{0.558}            & \multicolumn{1}{c|}{0.543}               & \multicolumn{1}{c|}{0.545}        & 0.547            \\ \cline{2-7} 
                                    & Dx                           & \multicolumn{1}{c|}{0.689}                & \multicolumn{1}{c|}{0.685}            & \multicolumn{1}{c|}{0.656}               & \multicolumn{1}{c|}{0.657}        & 0.660            \\ \hline
\multirow{7}{*}{\textbf{MIMIC-IV}}  & Mort                         & \multicolumn{1}{c|}{0.315}                & \multicolumn{1}{c|}{0.307}            & \multicolumn{1}{c|}{0.294}               & \multicolumn{1}{c|}{0.296}        & 0.294            \\ \cline{2-7} 
                                    & LOS3                         & \multicolumn{1}{c|}{0.648}                & \multicolumn{1}{c|}{0.644}            & \multicolumn{1}{c|}{0.609}               & \multicolumn{1}{c|}{0.613}        & 0.614            \\ \cline{2-7} 
                                    & LOS7                         & \multicolumn{1}{c|}{0.328}                & \multicolumn{1}{c|}{0.323}            & \multicolumn{1}{c|}{0.317}               & \multicolumn{1}{c|}{0.315}        & 0.316            \\ \cline{2-7} 
                                    & Readm                        & \multicolumn{1}{c|}{0.118}                & \multicolumn{1}{c|}{0.112}            & \multicolumn{1}{c|}{0.120}               & \multicolumn{1}{c|}{0.119}        & 0.120            \\ \cline{2-7} 
                                    & Fi\_ac                       & \multicolumn{1}{c|}{0.724}                & \multicolumn{1}{c|}{0.722}            & \multicolumn{1}{c|}{0.714}               & \multicolumn{1}{c|}{0.717}        & 0.719            \\ \cline{2-7} 
                                    & Im\_disch                    & \multicolumn{1}{c|}{0.412}                & \multicolumn{1}{c|}{0.410}            & \multicolumn{1}{c|}{0.368}               & \multicolumn{1}{c|}{0.373}        & 0.372            \\ \cline{2-7} 
                                    & Dx                           & \multicolumn{1}{c|}{0.834}                & \multicolumn{1}{c|}{0.836}            & \multicolumn{1}{c|}{0.816}               & \multicolumn{1}{c|}{0.817}        & 0.817            \\ \hline
\end{tabular}

\end{table}
The result of applying pretraining to our framework is in Table \ref{tab:pretrain}
For pre-training, fine-tuning is performed after pre-training with the entire input dataset except for the test set. In the context of conventional pre-training and transfer learning, transfer learning from a large hospital to a small hospital can be considered.
However, we only checked whether learning from pre-training gives benefits to the model or not compared to random initialization of model parameters, rather than fine-tuning on partial datasets after pre-training on the entire dataset. The experiment on the transfer situation from a large hospital to a small hospital is left as future work.

In DescEmb with a hierarchical structure, pretraining text within each event with MLM was performed in the event encoder part, but no significant performance improvement was seen. So, we proceeded with pretraining in the structure of flatten where we expect events can be seen by each other, rather than just learning text within the same event.

SPAN MLM is intended to learn the context of the EHR time series event by learning the event itself, rather than simply learning the partial random masked subword of the description.
The MLM accuracy of random masking is more than 90\%, but the accuracy of span MLM is about 80\%, resulting in a more difficult task for the model.
We haven’t seen any performance improvement with pre-training yet. A pre-training method suitable for the characteristics of EHR is needed to be newly developed.

\section{Qualitative analysis}\label{apd:qualitative}
\begin{table*}[ht]
    \caption{\label{tab: topfeatures} \textbf{Top 15 important features in mortality prediction of \oursarc{} trained on the pooled dataset (MIMIC-III+eICU).
    We accumulated the gradients for each event at the event encoder $f$, and ranked them in descending order.}}
    \centering
    \adjustbox{max width=0.7\textwidth}{
    \begin{tabular}{l l}
    \toprule
    MIMIC-III  &  eICU  \\
    \hline
    \midrule
    alendronate sodium po & anf / ana \\
    oxycodone sustained release po & d5w c bicarb \\
    morphine sulfate oral soln. po & pantoprazole protonix \\
    furosemide lasix 500 / 100 & vancomycin in ivpb \\
    acetaminophen - iv & nss w / versed / fent \\
    vancomycin hcl & rocuronium iv \\
    Norepinephrine & cisatracurium \\
    alpha - fetoprotein & oxycodone-acetaminophen 325mg \\
    pentamidine isethionate iv & vitamin d oral \\
    muItivitamin -12 i v & morphine 250 mg sodium chlorid \\
    heparin fIush port 10units/mI & norepinephrine bitartrate \\
    heparin fIush 5000 units/mI & rocuronium ivf infused \\
    ceftazidime & famotidine pepcid iv push \\
    acetaminophen - iv & 3 \% sodium chloride ivf infused \\
    timolol maleate 0. 25  & docusate sodium per ng tube \\
    ranitidine prophylaxis & amiodarone bolus ivpb \\
    \bottomrule
    \end{tabular}
    }
\end{table*}
\begin{table}[ht]
\centering
\caption{\label{tab:qual} Terms used for qualitative analysis}
\scriptsize
\begin{tabular}{|l|l|l|}
\hline
\multicolumn{1}{|c|}{\textbf{Term}} & \multicolumn{1}{c|}{\textbf{MIMIC-III}} & \multicolumn{1}{c|}{\textbf{eICU}} \\ \hline
vancomycin                          & hcl                                     & in ivpb                            \\ \hline
morphine                            & sulfate oral soln. po                   & 250 mg sodium chloride             \\ \hline
norepinephrine                      & -                                       & bitartrate iv                      \\ \hline
acetaminophen                       & iv                                      & oxycodone 325 mg po tabs           \\ \hline
\end{tabular}

\end{table}
We have seen so far that \oursarc{} was able to demonstrate quantitatively superior, or at least comparable predictive performance to all baselines for multiple prediction tasks, three EHR datasets, and three learning scenarios.

In this section, we provide a qualitative case study to see that \oursarc{} is not only reporting good AUPRC numbers, but it is also learning actually meaningful medical knowledge. 
In order to see which features were significant in predictive tasks, we accumulated the gradient of back-propagation of each event at the event encoder $f$. We followed the feature importance calculation method of DescEmb~\cite{hur2022unifying} on the mortality prediction with MIMIC-III and eICU.
 
We hypothesize that the larger the gradient, the more impactful the features.
The gradients for each event were tallied by the main feature of its corresponding event type (\textit{e.g.}, lab test name in the lab test event, or drug name in the prescription event).
We analyzed the top 100 important features in MIMIC-III and eICU, where the top 15 important features are provided in Table~\ref{tab: topfeatures} in descending order.
Within the top 100 features, we examined the features shared by both DescEmb* and \oursarc{} to show that \oursarc{} still utilizes meaningful features even without a careful feature selection process.

As a result, it turns out that both models share 87 and 79 out of the top 100 features in MIMIC-III and eICU, respectively, which means that \oursarc{} can figure out which features are significant for the predictive tasks without explicit guidance from human experts. The top 15 important features are described in Table \ref{tab: topfeatures}

Next, to test if UniHPF can handle the discrepancy between MIMIC-III and eICU in terms of textual description, we select four drug terms from the top 15 features  that exist in both datasets, and swap a part of terms between the two datasets, where the selected terms are described in Table~\ref{tab:qual}. For example, we switch all existing drugs vancomycin hcl'' in the test set of MIMIC-III to vancomycin in ivpb''.Then, we evaluate our model that was trained on each single dataset for mortality prediction, using the modified test set of MIMIC-III and eICU, respectively.

As a result, the AUPRC decreased marginally (0.8\%p and 0.6\%p in MIMIC-III and eICU, respectively) although the model never saw the modified features before (e.g., "vancomycin" in ivpb if the model has been trained on MIMIC-III). We conclude that UniHPF is able to deal with distinct EHR datasets, as long as they are based on the same language.
\end{document}